\documentclass[letterpaper, 10 pt, journal, twoside]{IEEEtran}

\usepackage{amsmath,amsfonts,amssymb}
\usepackage{array}
\usepackage{textcomp}
\usepackage{stfloats}
\usepackage{url}
\usepackage{graphicx}
\usepackage{siunitx}
\usepackage{bm}
\usepackage{cite}
\usepackage{subcaption}
\usepackage{balance}

\makeatletter
\let\NAT@parse\undefined
\makeatother
\usepackage[
	pdftitle={},
	pdfsubject={},
	pdfauthor={},
	pdfkeywords={},
	hidelinks
]{hyperref}

\newif\ifhighlightchanges
\ifhighlightchanges
	\usepackage{xcolor}
	\newcommand{\hlchanges}[1]{\textcolor{red}{#1}}
\else
	\newcommand{\hlchanges}[1]{#1}
\fi

\newif\ifarxiv
\arxivtrue   

\title{Learning Visually Interpretable Oscillator Networks for Soft Continuum Robots from Video}

\author{Henrik Krauss$^{1}$, Johann Licher$^{2}$, Naoya Takeishi$^{3}$, Annika Raatz$^{2}$, and Takehisa Yairi$^{3}$%
\thanks{Manuscript received: November 20, 2025; Revised: April 10, 2026; Accepted: June 3, 2026.}%
\thanks{This paper was recommended for publication by Editor Y.-L. Park upon evaluation of the Associate Editor and Reviewers' comments. This work was supported in part by JST PRESTO under Grant JPMJPR24T6, in part by JSPS KAKENHI under Grant JP25H01454, and in part by the Deutsche Forschungsgemeinschaft (DFG, German Research Foundation) under Grant No. 405030609.}%
\thanks{$^{1}$Henrik Krauss is with the Department of Advanced Interdisciplinary Studies, The University of Tokyo, Tokyo 153-8904, Japan {\tt\footnotesize henrik1.krauss@gmail.com}.}%
\thanks{$^{2}$Johann Licher and Annika Raatz are with the Institute of Assembly Technology and Robotics, Leibniz University Hannover, 30823 Hannover, Germany {\tt\footnotesize licher@match.uni-hannover.de}.}%
\thanks{$^{3}$Naoya Takeishi and Takehisa Yairi are with the Research Center for Advanced Science and Technology, The University of Tokyo, Tokyo 153-8904, Japan.}%
\ifarxiv
\thanks{Digital Object Identifier (DOI): 10.1109/LRA.2026.3703241}
\else
\thanks{Digital Object Identifier (DOI): see top of this page.}
\fi}

\begin{document}

\ifarxiv
\thispagestyle{empty}
\pagestyle{empty}
{\LARGE IEEE Copyright Notice}
\newline
\fboxrule=0.4pt \fboxsep=3pt

\fbox{\begin{minipage}{1.1\linewidth}
		\textcopyright~2026 IEEE. Personal use of this material is permitted. Permission from IEEE must be obtained for all other uses, in any current or future media, including reprinting/republishing this material for advertising or promotional purposes, creating new collective works, for resale or redistribution to servers or lists, or reuse of any copyrighted component of this work in other works.

		\vspace{0.5em}
		Accepted to be published in: IEEE Robotics and Automation Letters (RA-L), 2026.
		\vspace{0.5em}

        DOI: 10.1109/LRA.2026.3703241
\end{minipage}}
\fi

\maketitle

\begin{abstract}
\hlchanges{Learning soft continuum robot (SCR) dynamics from video offers flexibility but existing methods lack interpretability or rely on prior assumptions. Model-based approaches require prior knowledge and manual design.} We bridge this gap by introducing: (1) The Attention Broadcast Decoder (ABCD), a plug-and-play module for autoencoder-based latent dynamics learning that generates pixel-accurate attention maps localizing each latent dimension's contribution while filtering static backgrounds, \hlchanges{enabling visual interpretability via spatially grounded latents and on-image overlays.} (2) \hlchanges{Visual Oscillator Networks (VONs), a 2D latent oscillator network coupled to ABCD attention maps for on-image visualization of learned masses, coupling stiffness, and forces, thereby enabling mechanical interpretability.} We validate our approach on single- and double-segment SCRs, demonstrating that ABCD-based models significantly improve multi-step prediction accuracy with 5.8× error reduction for Koopman operators and 3.5× for oscillator networks on a two-segment robot. \hlchanges{VONs autonomously discover a chain structure of oscillators.} This fully data-driven approach yields compact, \hlchanges{mechanically interpretable} models \hlchanges{with potential relevance for} future control applications.
\end{abstract}


\section{Introduction}
\begin{figure}[t]
    \centering
    \includegraphics[width=\linewidth, trim=0 0mm 0 0, clip]{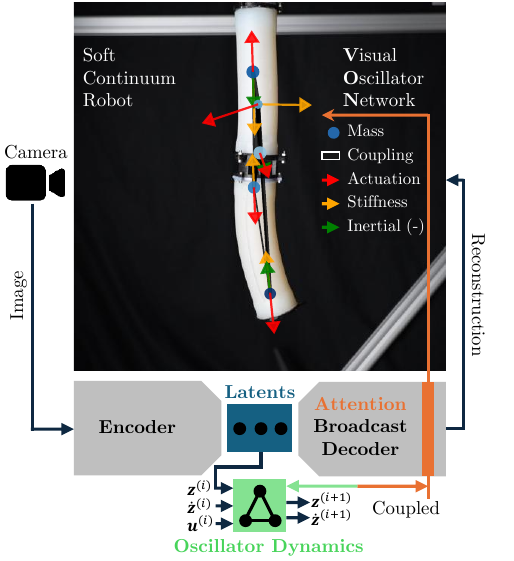}

    \caption{Main contributions of this study: (1) A newly proposed attention broadcast decoder (ABCD) is used plug-and-play for autoencoder-based Koopman and oscillator dynamics learning of a soft continuum robot. (2) Attention maps inside the ABCD are coupled to a 2D latent oscillators network to achieve VONs, so that masses, coupling stiffness, \hlchanges{stiffness forces, inertial forces,} and actuation forces can be visualized on the original, reconstructed, or predicted future images.}
    \label{fig:intro}
\end{figure}

\IEEEPARstart{T}{he} ability to learn dynamical systems directly from high-dimensional observations has recently transformed how we model and control complex robots. Koopman operator theory and deep learning-based embeddings enable data-driven discovery of latent dynamics without explicit modeling~\cite{brunton2022modern, lusch2018deep, li2020learning, takeishi_learning_2017}. Koopman operator methods provide a linear representation of nonlinear dynamics, facilitating prediction and control using linear methods~\cite{brunton2022modern}.

Using (coupled) oscillator networks to describe latent dynamics is a versatile low-order time sequence modeling approach with interpretability potential~\cite{lanthaler2023neural}. Learning dynamics directly from video data requires minimal setup and enables end-to-end learning. Building on this idea, Stölzle and Della Santina showed that input-to-state stable, coupled oscillator networks (CONs) can learn latent-space dynamics of SCRs from simulated video and support model-based control~\cite{stoelzle2024input}.

SCRs are an especially challenging research platform due to their infinite-dimensional, highly nonlinear mechanics~\cite{laschi_learningbased_2023}. Even if accurate models exist, they are often too computationally expensive for real-time control. As a result, data-driven dynamics learning has become an increasingly important direction in soft robotics. Recent reviews highlight the growing trend toward learning-based dynamics models for robotic manipulation~\cite{ai2025review}.

Overall, a trade-off exists between interpretability, performance, and modeling accuracy. Data-driven methods provide flexibility and simplicity but often lack interpretability and physical meaning. Model-driven approaches are interpretable and physically meaningful but require extensive prior modeling and are computationally heavy. Hybrid methods offer a promising compromise but still rely on manually designed physics components. To the best of our knowledge, no existing approach is fully data-driven yet yields a compact, \hlchanges{mechanically interpretable} low-parameter model suitable for control of SCRs.

In this work we address this gap by
\begin{enumerate}
    \item proposing a novel \emph{Attention BroadCast Decoder} (ABCD) for autoencoder-based latent dynamics learning from image sequences (video). \hlchanges{This enables \emph{visual interpretability} by assigning pixel-accurate attention maps for each latent, indicating its location in the image while filtering out the static background.}
    \item demonstrating how the attention maps produced by the ABCD can be coupled with a latent oscillator network to form \hlchanges{\emph{Visual Oscillator Networks (VONs)}, enabling on-image depiction of explicit oscillator models whose masses, coupling stiffnesses, and forces are \emph{mechanically interpretable}, in that their configuration and resulting motion corresponds to the real system.} This achieves unprecedented levels of interpretability of learned dynamics and identification of low-order dynamics without prior knowledge.
\end{enumerate}
This constitutes a novel approach for deriving an SCR model from video data without prior knowledge, exhibiting a level of \hlchanges{mechanical structure and interpretability} similar to manual models.

\section{Related Work}

\subsection{Hybrid and Data-Driven Modeling of SCRs}
Koopman-based models can lift the nonlinear dynamics of SCRs to a linear control space enabling real-time model predictive control (MPC)~\cite{bruder_datadriven_2021, shi2023koopman}. Haggerty et al.~\cite{haggerty2023control} showed that Koopman methods can even be used to control highly dynamic motions of SCRs. Ristich et al.~\cite{ristich_physicsinformed_2025b} introduced a hybrid Koopman approach that improves simulation accuracy from small datasets by additionally using phase-space information from a Cosserat rod model.

Physics-informed neural networks (PINNs) combine physical priors and data for highly flexible yet structured learning. Liu et al.~\cite{liu_physicsinformed_2024} use Lagrangian and Hamiltonian neural networks to control a tendon-driven SCR in simulation. However, real-robot results remain limited to quasi-static predictions due to the simplified underlying physics model. Licher et al.~\cite{licher2025adaptive} train a domain-decoupled PINN~\cite{krauss_domaindecoupled_2024b} on a full Cosserat rod model and utilize it for adaptive MPC and state estimation.

Reduced-order modeling (ROM) methods also play a significant role in SCR modeling. Alkayas et al.~\cite{alkayas2025structure} propose a structure-preserving autoencoder for fast simulation and control, with related work on optimal strain parameterization~\cite{alkayas2025soft}. 

Each of these methods has limitations: Koopman-based models do not offer a direct mechanical interpretation, PINNs require manual derivation of physics models, and ROMs depend on accurate high-fidelity simulations that must also be derived by hand.
\subsection{Vision-Based and Image-Driven Modeling of Soft Robots}

\hlchanges{Recent progress in vision-based learning highlights the potential of image-only sensing for soft robots. Zheng et al.~\cite{zheng2024vision} estimate key points for SCRs from stereo images, Rong and Gu~\cite{rong2024vision} perform real-time shape estimation under self-occlusion, and Almanzor et al.~\cite{almanzor_static_2023} learn deep visual inverse kinematics for static shape control.}

To also consider dynamics, Monteiro et al.~\cite{monteiro2024visuo} developed a visuo-dynamic self-modeling approach using recurrent networks, which learns from actuation inputs and video data to predict and control the motion of soft robots. A more interpretable hybrid model is presented by Valadas et al.~\cite{valadas_learning_2025a}, who learn a piecewise-constant strain model from video data and integrate it into a model-based controller.

\hlchanges{None of these vision-based methods yields a compact, mechanically interpretable dynamical model \emph{without prior system assumptions} (e.g., an explicit strain/curvature model and shape extraction). In contrast, our method learns dynamics end-to-end from raw images and provides visual and mechanical interpretability via attention-localized latents and VON visualizations.} The approach of Stölzle and Della Santina~\cite{stoelzle2024input} comes closest, yet their oscillators are not linked to the robot's physical configuration or structural elements, and the method is only demonstrated on simulated SCRs.

\section{Methods} \label{sec:methods}
\subsection{Autoencoder-Based Latent Dynamics Learning}
We implement an autoencoder-based latent dynamics learning approach for either Koopman or oscillator networks based on and modified from~\cite{lusch2018deep, stoelzle2024input}.

Given an input image $\bm{o}^{(i)} \in \mathbb{R}^{c \times h \times w}$ at time step $i$, we extract the latent coordinate
\begin{equation}
    \bm{z}^{(i)} = \varphi(\bm{o}^{(i)}) \in \mathbb{R}^{k}
\end{equation}
with the encoder network $\varphi$. \hlchanges{The latent state is defined as $\bm{\xi}^{(i)} = [\bm{z}^{(i)\top}, \bm{\dot{z}}^{(i)\top}]^\top \in \mathbb{R}^{2k}$.} We then predict the next latent state
\begin{equation}
    \hlchanges{
    \hat{\bm{\xi}}^{(i+1)} = f_\text{dyn}(\bm{\xi}^{(i)}, \bm{u}^{(i)})}
\end{equation}
after time interval $\Delta t$ with the dynamical model $f_\text{dyn}$. \hlchanges{The predicted latent coordinate $\hat{\bm{z}}^{(i+1)}$ is given by the first $k$ components of $\hat{\bm{\xi}}^{(i+1)}$.} With $\varphi$ and $\varphi^{-1}$ being approximated by neural networks, we can then retrieve the predicted next image
\begin{equation}
    \hlchanges{\hat{\bm{o}}^{(i+1)} = \varphi^{-1}(\hat{\bm{z}}^{(i+1)})}
\end{equation}
with the decoder network $\varphi^{-1}$. This approach is similar to~\cite{lusch2018deep}.

The latent velocities $\bm{\dot{z}}^{(i)}$ are obtained by mapping observation-space velocities to latent space via the encoder's Jacobian after~\cite{stoelzle2024input}.
Observation velocities are approximated using central finite differences which are then mapped to latent velocities using forward-mode automatic differentiation~(AD)
\begin{equation}
    \bm{\dot{z}}^{(i)} = \frac{\partial \varphi}{\partial \bm{o}}(\bm{o}^{(i)}) \cdot \bm{\dot{o}}^{(i)}
\end{equation}
where $\partial \varphi / \partial \bm{o}$ is the Jacobian of the encoder network evaluated at $\bm{o}^{(i)}$.

We use a $\beta$-VAE~\cite{higgins2017beta} for the encoder and decoder networks. Following~\cite{lusch2018deep}, the model is trained end-to-end with the basic loss
\begin{equation}
\begin{aligned}
\mathcal{L}_\text{basic} &= 
\frac{1}{N}\sum_{i=1}^N \Bigg[
    \underbrace{\text{MSE}(\varphi^{-1}(\bm{z}^{(i)}), \bm{o}^{(i)})}_{\text{static reconstruction}} \\
    &\quad + \lambda_\text{d} \, \underbrace{\text{MSE}(\varphi^{-1}(\hat{\bm{z}}^{(i+1)}), \bm{o}^{(i+1)})}_{\text{dynamic reconstruction}} \\
    &\quad + \beta \, \underbrace{\Big(-\frac{1}{2} \sum_j (1 + \log (\sigma_j^{(i)})^2 - (\mu_j^{(i)})^2 - (\sigma_j^{(i)})^2) \Big)}_{\text{KL divergence}} \\
    &\quad + \lambda_z \, 
        \underbrace{
        \begin{aligned}
            &\Big(\text{MSE}(\hat{\bm{z}}^{(i+1)}, \bm{z}^{(i+1)}) \\
            &+ \text{MSE}(\Delta t \cdot \hat{\bm{\dot{z}}}^{(i+1)}, \Delta t \cdot \bm{\dot{z}}^{(i+1)}) \Big)
        \end{aligned}
        }_{\text{latent dynamics consistency}}
\Bigg],
\end{aligned}
\end{equation}
for a batch of size $N$, where $\sigma_j^{(i)}$ and $\mu_j^{(i)}$ are the standard deviation and mean of the $j$-th latent dimension $z_j^{(i)}$ of $\bm{z}^{(i)}$. Additional losses as described below and in Sec.~\ref{subsec:ABCD} are added inside $\sum_i^N$ depending on network configuration.

For $f_\text{dyn}$, we employ both Koopman operators and oscillator networks:

\textit{Koopman dynamics:}
We learn a Koopman operator on the latent state $\bm{\xi}^{(i)} = [\bm{z}^{(i)\top}, \bm{\dot{z}}^{(i)\top}]^\top$ with learnable transition matrix $\bm{A} \in \mathbb{R}^{2 k \times 2 k}$ and feed-forward neural network (FNN) $\bm{B}(\cdot)$. The update is given by
\begin{equation}
    \hlchanges{\hat{\bm{\xi}}^{(i+1)} = \bm{A} \bm{\xi}^{(i)} + \bm{B}(\bm{u}^{(i)}).}
\end{equation}

\textit{Oscillator network:}
We use learnable positive-definite mass $\bm{M}$ (diagonal), stiffness $\bm{K}$, and Rayleigh damping matrices $\bm{D} = \alpha \bm{M} + \beta \bm{K}$ with learnable $\alpha, \beta \ge 0$. The dynamics are governed by
\begin{equation}
    \bm{M} \ddot{\bm{z}}^{(i)} + \bm{D} \dot{\bm{z}}^{(i)} + \bm{K} (\bm{z}^{(i)} - \bm{z}_0) = \bm{B}(\bm{u}^{(i)}).
\end{equation}
\hlchanges{Here, $\bm{z}_0$ is a learned equilibrium (rest) latent coordinate and can be non-zero.}
Symplectic Euler integration is used during training for stability. \hlchanges{In our implementation, the control input $\bm{u}^{(i)}$ is augmented with four delayed actuation inputs.} For oscillator networks, we augment $\mathcal{L}_\text{basic}$ with a \hlchanges{rest-state} loss $\lambda_\text{r} \mathcal{L}_\text{r}$ to enforce $\bm{z}^{(i)} = \bm{z}_0$ and $\bm{\dot{z}}^{(i)} = \bm{0}$ at equilibrium, encouraging physically consistent \hlchanges{rest-state} dynamics.

\subsection{Attention Broadcast Decoder}
\label{subsec:ABCD}
\begin{figure}[t]
    \centering
    \includegraphics[width=\linewidth, trim=0 10 0 0, clip]{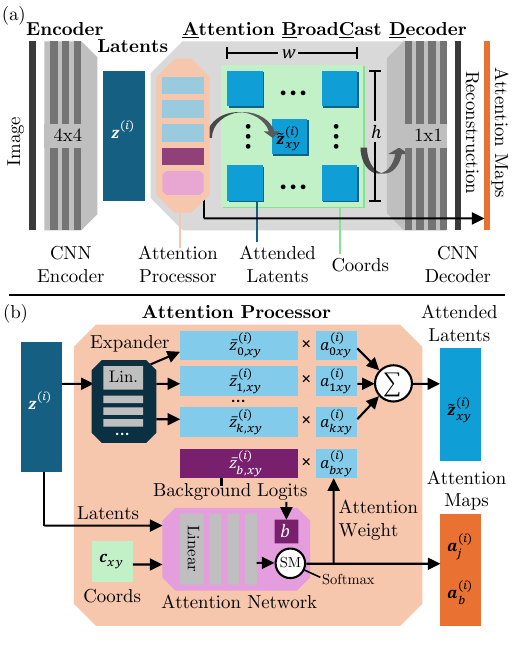}

    \caption{(a) The attention broadcast decoder (ABCD) integrated plug-and-play in an autoencoder setup for image-reconstruction learning. (b) The attention processor within the ABCD that retrieves attention maps and attended latents before they are spatially broadcasted and decoded.}
    \label{fig:architecture}
\end{figure}
\subsubsection{Architecture}
\label{subsec:ABCD_architecture}
We introduce the Attention Broadcast Decoder (ABCD) as a plug-and-play module for autoencoder-based latent dynamics learning.
It is inspired by the spatial broadcast decoder of Watters et al.~\cite{watters2019spatial}. As shown in Fig.~\ref{fig:architecture}(a), the ABCD takes latent coordinates $\bm{z}^{(i)}$ and produces reconstructed images and pixel-accurate attention maps. While the ABCD replaces a standard convolutional neural network (CNN) decoder, it implicitly affects the latent-state structure and the learned dynamics.

The attention processor (Fig.~\ref{fig:architecture}(b)) computes attended latents as follows:
An FNN $f_\text{att}$ produces attention logits
\begin{equation}
\ell_j^{(i)}(x,y) = f_\text{att}([\bm{z}^{(i)}, \bm{c}_{xy}])_j
\end{equation}
evaluating the latent coordinate $\bm{z}^{(i)}$ at each coordinate pair $\bm{c}_{xy} \in [-1,1]^2$.
Pixel-wise attention values at location $(x, y)$
\begin{equation}
    a_j^{(i)}(x,y) = \text{softmax}([\ell_1^{(i)}(x,y), \ldots, \ell_k^{(i)}(x,y), b])_j,
\end{equation}
are calculated by concatenating the logits with a constant background logit $b \equiv \ell^{(i)}_{k+1}(x,y)$ (a hyperparameter) and normalizing via softmax, resulting in attention maps $\bm{a}_j^{(i)} \in \mathbb{R}^{h \times w}$ for $j=1,\ldots,k+1$, including the background attention map $\bm{a}_b^{(i)}$. Separately, each scalar latent $z_j^{(i)}$ is expanded to extended latents 
\begin{equation}
    \overline{\bm{z}}_{j,xy}^{(i)} = \bm{W}_j z_j^{(i)} + \bm{b}_j \in \mathbb{R}^{n_f},
\end{equation}
via a linear layer of parameters $\bm{W}_j$ and $\bm{b}_j$ and broadcasted to image resolution. Additionally, learnable spatial background features $\overline{\bm{z}}_b$ provide an auxiliary feature map for capturing static background content. The attention-weighted latent coordinate maps are then computed by
\begin{equation}
    \tilde{\bm{z}}^{(i)} = \sum_{j=1}^{k} \bm{a}_j^{(i)} \odot \overline{\bm{z}}_{j}^{(i)} + \bm{a}_b^{(i)} \odot \overline{\bm{z}}_b.
\end{equation}
where $\odot$ denotes element-wise multiplication.
These are concatenated with coordinates and processed by four $1{\times}1$ CNN layers to construct the output image. Unlike spatial broadcast decoders~\cite{watters2019spatial}, the ABCD makes individual latents compete for reconstructing each pixel via the attention maps and decouples static background reconstruction from the latent state via learnable background features.

\subsubsection{Attention consistency loss}
While the ABCD can be used as described in Sec.~\ref{subsec:ABCD_architecture}, we introduce an additional \emph{attention consistency loss}
\begin{equation}
    \mathcal{L}_\text{attn-cons} =
        \frac{1}{khw} \sum_{j=1}^k \sum_{x,y} 
            \left| \dot{a}_j^{(i)}(x, y) \right| \cdot \left(1 - \dot{\bar{\bm{o}}}^{(i)}(x, y) \right)
    ,
\end{equation}
to regularize the attention dynamics, where $\dot{a}_j^{(i)}(x, y)$ is the time derivative of the attention map for latent $j$ at spatial location $(x, y)$ and $\dot{\bar{\bm{o}}}^{(i)}(x, y)$ is the normalized observation velocity at pixel $(x, y)$. This loss penalizes attention changes at pixels without image motion, encouraging sharper separation between background and dynamic regions.

\subsubsection{\hlchanges{Visual Oscillator Networks (VONs)}}
\hlchanges{For VONs,} we introduce an additional modification that interprets the learned dynamics as a \hlchanges{VON} coupled to the attention maps to enable on-image visualization. We group consecutive pairs of latent dimensions into $n = k/2$ two-dimensional oscillators, where oscillator $l \in \{1,\ldots,n\}$ has latent-space position $\bm{q}_l^{(i)} = [z_{2l-1}^{(i)}, z_{2l}^{(i)}]^\top \in \mathbb{R}^2$ and velocity $\dot{\bm{q}}_l^{(i)} = [\dot{z}_{2l-1}^{(i)}, \dot{z}_{2l}^{(i)}]^\top \in \mathbb{R}^2$. We modify the ABCD's attention processor to generate one attention map $\bm{a}_l^{(i)} \in \mathbb{R}^{h \times w}$ per oscillator (instead of per individual latent dimension). Correspondingly, the mass matrix $\bm{M}$ is constrained such that $M_{2l-1,2l-1} = M_{2l,2l}$ for all $l$, ensuring equal mass for both dimensions of each 2D oscillator, while the damping and stiffness matrices $\bm{D}$ and $\bm{K}$ remain unconstrained. To identify oscillator positions into image space, we compute the center-of-mass (COM) of the squared attention maps:
\begin{equation}
    \bm{p}_l^{(i)} = \frac{\sum_{x,y} (a_l^{(i)}(x,y))^2 \cdot \bm{c}_{xy}}{\sum_{x,y} (a_l^{(i)}(x,y))^2} \in [-1,1]^2,
\end{equation}
where $\bm{c}_{xy} = [x, y]^\top$ are normalized pixel coordinates and squaring emphasizes regions of low mutual attention overlap, acting as a soft indicator of individual oscillator localization. The COM velocity $\dot{\bm{p}}_l^{(i)}$ is computed via the quotient rule by propagating $\bm{\dot{o}}^{(i)}$ through the attention mechanism using forward-mode AD. To ensure consistency between latent-space and image-space dynamics, we introduce an \emph{attention coupling loss} that enforces matching pairwise relative motions:
\begin{equation}
    \mathcal{L}_\text{attn-coupling} = \mathbb{E}_{l,m\neq l} \left[ \left( \frac{\dot{d}_{lm}^\text{lat}(i)}{\bar{v}_{lm}^\text{lat}(i)} - \frac{\dot{d}_{lm}^\text{img}(i)}{\bar{v}_{lm}^\text{img}(i)} \right)^2 \right],
\end{equation}
where $\dot{d}_{lm}^\text{lat}(i)$ and $\dot{d}_{lm}^\text{img}(i)$ denote the signed relative velocities between oscillators $l$ and $m$ projected along their connecting direction in latent space ($\bm{q}_l, \bm{q}_m$) and image space ($\bm{p}_l, \bm{p}_m$), respectively, and $\bar{v}_{lm}(i)$ is the mean velocity magnitude of each pair. This approach assumes that distances in the image are proportional to the physical distances between the corresponding points on the robot.

\hlchanges{To visualize forces in image space after training, we map latent forces to the image plane using the local, state-dependent Jacobian $J_l^{(i)} = \frac{\partial p_l^{(i)}}{\partial q_l^{(i)}}$, which is analytically derived via automatic differentiation through the ABCD's latent-to-image mapping.}

\subsection{Soft Continuum Robot Datasets}
\hlchanges{We investigate one single-segment SCR and one two-segment SCR with a rigid connector (\SI{21}{mm} long).} The segments are soft pneumatic actuators casted from silicone with an outer diameter of \SI{42.4}{mm} and a length of \SI{130}{mm} and entail three fiber-reinforced chambers arranged at \SI{120}{\degree} intervals around an inner channel. \hlchanges{While the actuators can in principle move in 3D, we enforce planar motion observable by a single camera by pressure-coupling two chambers per actuator. Pressures are controlled at 1~kHz using a PID controller. Video is recorded at \SI{120}{fps} in Full HD.} \hlchanges{Further hardware and manufacturing details are provided in~\cite{bartholdt_parameter_2021}.}

To generate diverse motions covering the full workspace and a wide spectrum of frequencies, 75 pressure trajectories with randomly sampled frequencies from \SI{0.04}{Hz} to \SI{2}{Hz} and random phase shifts are generated. Each trajectory consists of two superposed frequencies. These \SI{10}{s}-long trajectories are concatenated with a \SI{2}{s}-long linear interpolated transition in between, which results in an overall duration of \SI{15}{min}. The linear interpolation reduces the excitation of natural oscillations, which would bury the controlled oscillation.

The dataset is made publicly available\footnote{Dataset available at: \url{https://zenodo.org/records/17812071}}.

\section{Results}\label{sec:results}

\subsection{Neural-Network Training Setup}
All models are implemented in PyTorch and four configurations for each dataset are trained: (1)~Koopman with standard decoder, (2)~Koopman with ABCD ($+\mathcal{L}_\text{attn-cons}$), (3)~1D oscillator network with standard decoder, and (4)~\hlchanges{VONs} ($+\mathcal{L}_\text{attn-cons}$, $+\mathcal{L}_\text{attn-coupling}$). Images are downscaled to $3 \times 32 \times 32$ pixels and subsampled to 60 fps ($\Delta t = 1/60$ s). 
\hlchanges{Based on an oscillator-count hyperparameter sensitivity study (see on Fig.~\ref{fig:multistep_graph} (right)), we use $n=3$ oscillators ($k=6$ latents) for the 1-segment dataset and $n=5$ oscillators ($k=10$ latents) for the 2-segment dataset. For fairness, the Koopman models use the same latent dimension $k$ as the corresponding VON setting.} All models use the same encoder architecture (3-layer CNN + linear layer) and the standard decoder mirrors it with transposed convolutions.
We train all models for 300 epochs using AdamW with a batch size of 32 and separate learning rates for encoder, dynamics, and decoder. For ABCD-based models, we use a short warmup phase with reduced encoder and dynamics learning rate to encourage background information to be captured by the learnable background parameters. Training is terminated early at 100 epochs for the \hlchanges{VONs}, as we observe better generalization. The implementation including the detailed architecture, hyperparameters, loss weights, and training schedules is available on GitHub\footnote{Code available at: \url{https://github.com/UThenrik/visual_oscillators_for_SCR}}.
A video containing animations of the results presented in Sec.~\ref{subsec:attn_maps}-\ref{subsec:multi_step_prediction} is provided as supplemental media and is also available on YouTube\footnote{Video available at: \url{https://youtu.be/i80H8erVISM}}.

\subsection{Attention Maps}
\label{subsec:attn_maps}
\begin{figure}[h]
     \centering
     \includegraphics[width=\linewidth, trim=0 10mm 0 0 0, clip]{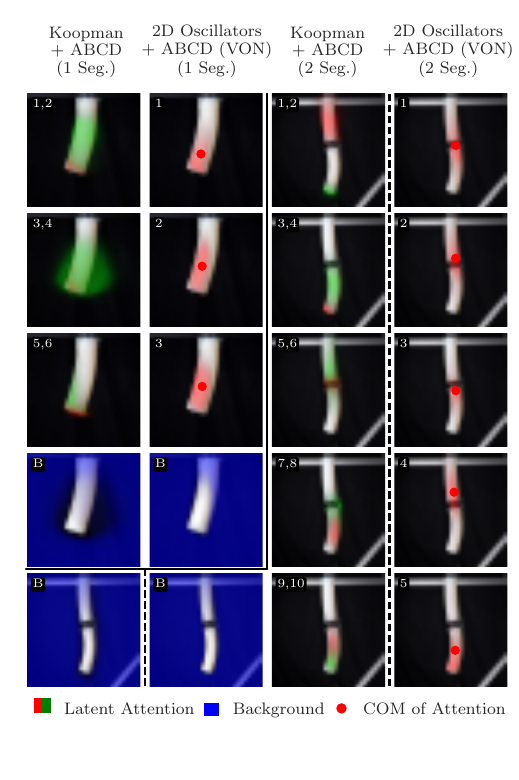}

    \caption{Attention maps for the 1-segment and 2-segment robots for Koopman and oscillator networks using the ABCD. Two attention maps are shown per image for the Koopman models for visual compactness.}
    \label{fig:attention_maps_vis}
\end{figure}

Fig.~\ref{fig:attention_maps_vis} shows the attention maps for the 1-segment and 2-segment robots for Koopman and oscillator networks using the ABCD, visualizing the relative contribution of each latent dimension to the reconstruction of each pixel or to the static background. We can see that all models identify the static background accurately, \hlchanges{with VON models showing sharper seperation.} A continuous gradient at the base of the SCR from latent attention maps to background map is observed.
The latent attention maps focus on different regions of the robots, showing the different information of the robot they encode. By visualizing the COM of the squared attention maps for the 2D oscillators \hlchanges{of the VONs}, we can also see that they are distributed on and along the main axis of the 1-segment and 2-segment SCR.

\subsection{\hlchanges{Visual Oscillator Networks (VONs)}}
\label{subsec:oscillator_nets}
\begin{figure*}[t]
    \centering
    \includegraphics[width=\linewidth, trim=0 10 0 0, clip]{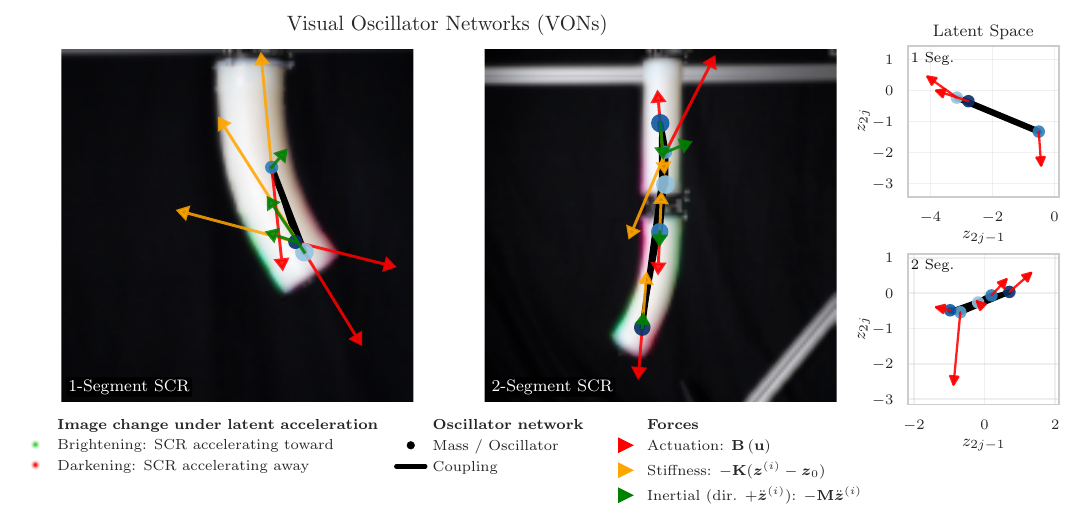}

    \caption{\hlchanges{VONs} for the 1-segment and 2-segment robots (left) and latent space visualization (right) on a validation state. The \hlchanges{VONs}, which accurately capture the SCR dynamics, are shown with their masses, coupling stiffness, \hlchanges{stiffness forces, inertial forces (in acceleration direction), and} current actuation forces. \hlchanges{Marker size, coupling line thickness, and force arrow lengths are proportional to the corresponding quantities.} \hlchanges{Coupling-to-ground stiffness and damping are learned but not visualized to avoid visual clutter (i.e., the learned oscillator system is not free-floating).} \hlchanges{Acceleration of the SCR is shown in image space by coloring the SCR edge based on image brightening and darkening from extrapolating and decoding the shown state based on latent accelerations.} Oscillator positions are scaled \hlchanges{1.75} times around their shared mean position and higher resolution images are used for visual clarity.}
    \label{fig:oscillator_network_vis}
\end{figure*}

Figure~\ref{fig:oscillator_network_vis} shows the learned \hlchanges{VONs} visualized on the corresponding robot images. Oscillator masses are represented by marker size, coupling stiffness by line width, \hlchanges{and actuation, stiffness, and inertial forces by arrows.} Markers are colored in different shades of blue to enable identification of the same oscillators in the corresponding latent space visualization shown on the right. The relative positions of the oscillator COMs in image space match their relative positioning when latents are viewed as 2D coordinates in latent space, as enforced by the attention coupling loss $\mathcal{L}_\text{attn-coupling}$. \hlchanges{Stiffness forces act as restoring forces toward the rest configuration, while actuation often acts in the opposite direction. Inertial forces indicate the resulting acceleration while damping forces are not shown to prevent visual clutter. SCR acceleration is additionally visualized on the image and reasonably matches the inertial force directions.}

\hlchanges{For the 1-segment robot, the actuation forces push the SCR to the lower right with stiffness forces pointing toward rest position. As the latter are stronger, the SCR's tip is accelerated to the left, as visible by the green edge. This well matches the inertial oscillator forces. For the 2-segment robot, the stiffness forces point at return-to-rest for the different oscillators, and inertial forces predict a more complex movement with the upper part moving right and the tip moving upward. This also reasonably well matches the real image acceleration denoted by the green edge.}

Notably, for the 2-segment robot the learned structure forms a chain of five oscillators, which is \hlchanges{a reasonable low-order representation}. In Cosserat rod theory, soft continuum robots can be viewed as an infinite chain of coupled oscillators, and the ABCD-based oscillator network identifies a finite chain as a reduced-order approximation. The spatial distribution of oscillators is also insightful, as two oscillators at each end are spread out along the robot, while three middle oscillators cluster near the connecting segment and exhibit high mutual stiffness. This structure aligns with the physical construction, as the two-segment robot is connected by a stiff intermediate element. \hlchanges{Like the real system, the chain elongates under actuation and contracts toward the rest state (see supplemental animation).}

\subsection{Single and Multi-Step Prediction}
\label{subsec:multi_step_prediction}
\begin{figure}[h!]
    \centering
    \includegraphics[width=\linewidth, trim=0 0 0 0, clip]{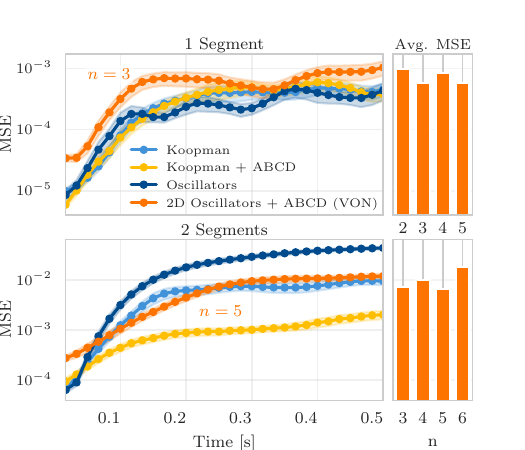}

    \caption{(Left) Multi-step  reconstruction error \hlchanges{image MSE} over \SI{0.5}{s} for 1-segment (top) and 2-segment (bottom) robots. Shaded regions indicate standard error of the mean over 50 validation trajectories. \hlchanges{(Right) Multi-step errors from a VON oscillator-count sensitivity study.} The ABCD improves multi-step reconstruction accuracy for the more complex 2-segment system for both Koopman and oscillator network dynamics.}
    \label{fig:multistep_graph}
\end{figure}
\begin{figure*}[t]
    \centering
    \includegraphics[width=\linewidth, trim=0 0mm 0 0, clip]{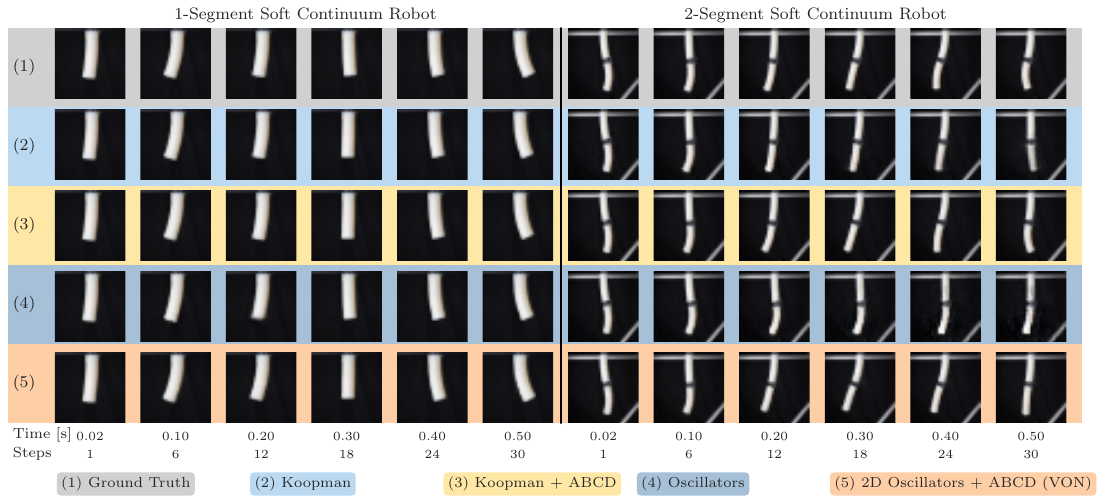}

    \caption{Multi-step reconstruction images (1\textsuperscript{st}, and every 6\textsuperscript{th} step shown) over \SI{0.5}{s} for 1- and 2-segment robots, compared to ground truth for a validation trajectory. All 1-segment models achieve highly accurate reconstructions. For 2 segments, the ABCD improves long-term reconstruction accuracy for both Koopman and oscillator network dynamics.}
    \label{fig:multistep_vis}
\end{figure*}
We evaluate all trained networks on single-step and multi-step prediction tasks on validation trajectories. We initialize each model with an observation and its latent velocity (computed via central differences and forward-mode AD), then autoregressively predict the next 30 steps (\SI{0.5}{s}) by repeatedly applying the learned dynamics. \hlchanges{All reported prediction errors are image MSEs between the predicted frame and the ground-truth observation.}

Fig.~\ref{fig:multistep_graph} shows the reconstruction error over time and steps for both 1-segment and 2-segment robots, and Fig.~\ref{fig:multistep_vis} shows the reconstruction images. \hlchanges{For the 1-segment robot, errors are similar across models, with multi-step MSE on the order of $5\cdot 10^{-4}$. The plain oscillator achieves $2.46\times10^{-4}$ and the VON achieves $5.74\times10^{-4}$.} \hlchanges{For the more complex 2-segment robot, Koopman+ABCD achieves the best multi-step MSE of $9.84 \times 10^{-4}$ compared to $5.66 \times 10^{-3}$ for standard Koopman, which corresponds to a $5.8\times$ improvement. The VON achieves $6.56 \times 10^{-3}$ compared to $2.27 \times 10^{-2}$ for the standard oscillator, which corresponds to a $3.5\times$ improvement.} \hlchanges{ABCD-based models have higher single-step image MSE, but lower long-horizon drift that improves multi-step prediction.}
\hlchanges{This trend is consistent for binarized-image MSE and intersection over union (IoU), indicating that it reflects shape prediction rather than image artifacts. This suggests strong generalization of the VONs and may be aided by the dynamical consistency encouraged by the attention losses and by the inductive bias of the ABCD. The reconstruction images visualized in Fig.~\ref{fig:multistep_vis} also show how the ABCD-based models more accurately predict the 2-segment SCR shape for higher multi-step counts, and that all models predict the 1-segment SCR shape accurately.}

\subsection{Latent-Space Extrapolation}
\begin{figure}[!t]
    \centering
    \includegraphics[width=\linewidth, trim=0 2mm 0 0, clip]{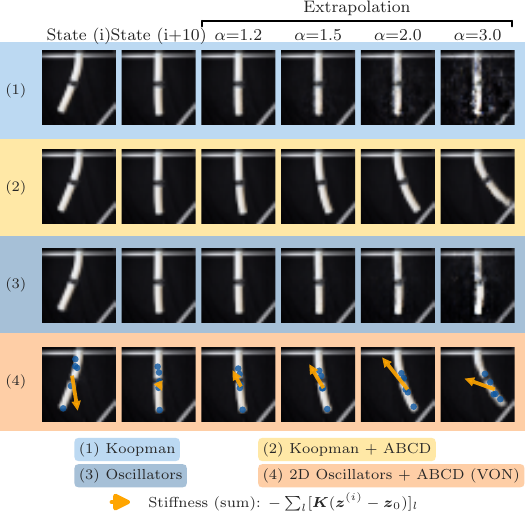}

    \caption{Latent-space extrapolation for 2-segment robot models. Two states (left-bent and near-rest) are linearly extrapolated in latent space with factor $\alpha$. Blue markers indicate oscillator positions of the VON. ABCD models successfully extrapolate a bending transitions from left to right, while standard models produce non-bent, noisy states. \hlchanges{For the VON, stiffness force sum is additionally shown, which acts as a restoring force toward equilibrium.}}
    \label{fig:extrapolation}
\end{figure}

We perform linear extrapolation by selecting two latent states from the data set, separated by 10 time steps, representing a transition of the robot from a left-bent configuration to an approximately neutral position. For extrapolation factors $\alpha \in \{1.2, 1.5, 2, 3\}$, we compute $\bm{z}_\text{extra} = \bm{z}^{(i)} + \alpha(\bm{z}^{(i+10)} - \bm{z}^{(i)})$ and decode the extrapolated latent states.

Fig.~\ref{fig:extrapolation} reveals a significant difference between models: Both Koopman with ABCD and the \hlchanges{VON} successfully extrapolate the bending movement, with a transition from the left-to-right bending motion, beyond a bending angle that is represented in the dataset (for $\alpha=3$). \hlchanges{The summed stiffness force of the VON remains a plausible restoring signal and increases for stronger extrapolations.} In contrast, the standard Koopman and oscillator networks without ABCD decoder fail to generate a meaningful extrapolation. Their reconstructions become trapped at the near-neutral configuration and produce noisy images. This indicates that the attention mechanism not only improves interpretability and multi-step prediction accuracy for the 2-segment SCR but also learns more structured latent representations with reasonable return-to-rest stiffness forces. \hlchanges{The physical realizability of these extrapolated states remains an open question, but the result may be relevant for future control applications and should be validated in dedicated control experiments.}

\section{Conclusions}\label{sec:conclusions}
\hlchanges{This work introduced the Attention Broadcast Decoder (ABCD) and \emph{Visual Oscillator Networks (VONs)} for autoencoder-based latent dynamics learning from video.} The ABCD generates pixel-precise attention maps for each latent dimension to enable their spatial localization, and separates static background from dynamic robot motion. We demonstrated its plug-and-play integration with both Koopman operators and oscillator networks for dynamics learning of a single- and dual-segment soft continuum robot (SCR).

\hlchanges{By coupling attention maps to VONs through an attention coupling loss, we enabled direct on-image visualization of learned dynamics, including masses, coupling stiffness, and forces, without prior knowledge of the system.} For the 2-segment robot, the model autonomously discovered a chain structure similar to Cosserat rod theory. The ABCD not only provides unprecedented \hlchanges{visual and mechanical interpretability} but also significantly improves multi-step prediction accuracy for the complex 2-segment robot, achieving \hlchanges{\SI{5.8}{\times} error reduction for Koopman and \SI{3.5}{\times} for oscillator networks.} Both Koopman and \hlchanges{VON} variants with ABCD further enable smooth latent space extrapolation where standard methods fail. \hlchanges{In our experiments, only 3 and 5 2D oscillators are required for the 1- and 2-segment robots, respectively, indicating computational efficiency.}

\hlchanges{This study has several limitations that also motivate future work. While the learned models may be well suited for future control applications due to their structured latent dynamics and long-horizon stability, dedicated control experiments are required. Further physical interpretability could be achieved through boundary conditions (e.g., a fixed base) or via additional constraints or regularization to better relate learned parameters to physical quantities. Strong latent-space extrapolation should be validated for physical realizability through dedicated out-of-distribution experiments or simulation. Augmenting the dataset with geometric ground truth (e.g., tip position) and evaluating under varying viewpoints would enable more direct geometric evaluation and broader generalization. \hlchanges{Our experiments further assume largely static backgrounds. Extending to variable backgrounds is a promising direction and may be integrated into the ABCD attention-based separation of robot and background.}}

\hlchanges{Beyond these limitations, promising further developments include applications to more diverse soft robots, multi-camera setups, and learning 3D oscillator networks akin to finite element models, as well as studying transfer to different camera perspectives.}

\balance
\bibliographystyle{IEEEtran}
\bibliography{references}

\end{document}